\documentclass{article}

\PassOptionsToPackage{numbers, compress}{natbib}

\usepackage[preprint]{neurips_2019}

\usepackage[utf8]{inputenc} 
\usepackage[T1]{fontenc}    
\usepackage{hyperref}       
\usepackage{url}            
\usepackage{booktabs}       
\usepackage{amsfonts}       
\usepackage{nicefrac}       
\usepackage{microtype}      
\usepackage{amsmath}
\usepackage{amssymb}
\usepackage{multirow}
\usepackage{bm}
\usepackage{graphicx}
\usepackage{caption}
\usepackage{subcaption}
\usepackage{wrapfig}
\usepackage[shortlabels]{enumitem}
\usepackage[flushleft]{threeparttable} 

\title{Weight Normalization based Quantization \\ for Deep Neural Network Compression}

\author{%
  Wen-Pu Cai \\
  National Key Lab. for Novel Software Tech. \\
  Dept. of Comp. Sci. and Tech. \\
  Nanjing University, Nanjing 210023, China \\
  \texttt{caiwp@lamda.nju.edu.cn} \\
  \And
  Wu-Jun Li \\
  National Key Lab. for Novel Software Tech.\\ 
  Dept. of Comp. Sci. and Tech. \\
  Nanjing University, Nanjing 210023, China \\
  \texttt{liwujun@nju.edu.cn} \\
}

\begin{document}

\maketitle

\begin{abstract}
      With the development of deep neural networks, the size of network models becomes larger and larger. 
      Model compression has become an urgent need for deploying these network models to mobile or embedded devices. 
      Model quantization is a representative model compression technique. 
      Although a lot of quantization methods have been proposed, many of them suffer from a high quantization error caused by a long-tail distribution of network weights. 
      In this paper, we propose a novel quantization method, called \underline{w}eight \underline{n}ormalization based \underline{q}uantization~(WNQ), for model compression. 
      WNQ adopts weight normalization to avoid the long-tail distribution of network weights and subsequently reduces the quantization error. 
      Experiments on CIFAR-100 and ImageNet show that WNQ can outperform other baselines to achieve state-of-the-art performance.

\end{abstract}

\section{Introduction}

In recent years, deep neural networks~(DNNs) have become one of the most popular models in machine learning and related areas like computer vision. 
In general, DNN models have a large amount of weight parameters and require ultra-high computational effort, leading to huge storage and computation overhead. 
For example, ResNet152~\cite{ResNet} has sixty million parameters and eleven billion float point operations~(FLOPs).
Hence, it is typically hard to deploy these network models to mobile or embedded devices.

To alleviate the aforementioned problem, deep neural network compression~(also called model compression) has become an urgent need. 
Recent works show that a large amount of redundancy exists in DNNs~\cite{han}. 
Therefore, we can greatly reduce the number of network parameters without significant loss of accuracy if appropriate model compression methods are designed. 
Recently, more and more methods have been proposed to compress and accelerate DNNs. 
These model compression methods include pruning~\cite{han,Thinet}, quantization~\cite{BC,ABC,LQ-NET}, tensor decomposition~\cite{TT-net,Filter_Group} and knowledge distillation~\cite{Distill,Mutual}.

Quantization tries to reduce the precision of network weights from 32-bit to low bit-width, which can dramatically reduce the model size. 
Besides, with proper quantization, the convolution operations can be replaced with only addition operations~\cite{TWN}, which can enable fast inference on embedded devices. 
Hence, model quantization has become a representative model compression technique, and a lot of quantization methods have recently been proposed~\cite{BC,ABC,LQ-NET,Integer}.

Although some quantization methods can quantize network weight to 8-bit representation without accuracy drop~\cite{Integer}, 
the accuracy may drop dramatically when the bit-width decreases extremely~(to 2-bit or 3-bit). 
The main reason is that the quantization error typically increases when decreasing the bit-width. 
To preserve the accuracy of the full precision network, we need to design quantization methods which can reduce the quantization error as much as possible. 
As one of the state-of-the-art quantization methods, LQ-Net~\cite{LQ-NET} attempts to quantize the weights and activations by minimizing mean-squared quantization error. 
Although LQ-Net has achieved the best performance in many cases, in our experiment we find that the long-tail weight distribution may be an important factor to cause a high quantization error in LQ-Net. 
Actually, this phenomenon of long-tail weight distribution is common in DNNs~\cite{Truct_Gaussian}. 
Hence, one possible way for reducing quantization error is to avoid the long-tail in the weight distribution during quantization.

In this paper, we propose a novel quantization method, called \underline{w}eight \underline{n}ormalization based \underline{q}uantization~(WNQ), for model compression. 
WNQ adopts weight normalization to avoid the long-tail distribution of network weights, and subsequently reduces the quantization error. 
WNQ is simple but effective. Experiments on CIFAR-100 and ImageNet show that WNQ can outperform other baselines to achieve state-of-the-art performance. 
To the best of our knowledge, this is the first work to study the effect of weight distribution for quantization.

\section{Related Work}\label{sec:related}
Model quantization attempts to exploit low-precision~(low bit-width) approximation of the original model parameters. 
\citet{BC} propose weight binarization in neural networks. They calculate the gradient with the binary weights, but accumulate gradient updates in the float weights. 
\citet{TWN} expand binary weights to ternary weights to approximate float weights more accurately. 
In general, both binary and ternary weights cannot preserve the performance of the original model in most cases. 
To further reduce the gap between the full precision models and the low-precision models, more works~\cite{Dorefa, Residual,ABC,ADMM, LQ-NET} introduce multi-bit quantization. 
Among them, \citet{Dorefa} adopt an intuitive and simple quantization method, which does not consider the real distribution of weights. 
\citet{Residual} propose residual quantization in which the role of each bit is to binarize the residual quantization error from all previous bits. 
Residual quantization takes weight distribution into consideration. However, it cannot achieve the minimum quantization error. 
\citet{ADMM} utilize ADMM to deal with discrete weight optimization, but the training time is too long. 
\citet{LQ-NET} propose LQ-Net to minimize quantization error by alternating optimization. 
Although LQ-Net can get minimum quantization error given the weights, it suffers from long-tail distribution of weights, which will incur
large quantization error. 
To further accelerate the speed of inference, many methods~\cite{BNN,Xnor-net,Dorefa,HWGQ,LQ-NET,TEL,QIL} also quantize activations to low precision. In this paper, we only focus on weight quantization. Existing activation quantization methods can be easily integrated with our weight quantization method, which will be pursued in our future work.

Besides model quantization methods, pruning~\cite{han,Group_Lasso,Weight_Sum,Thinet,Strength}, tensor decomposition~\cite{Tucker,TT-net,Filter_Group} and knowledge distillation~\cite{Distill,Mutual} are also widely used techniques for model compression. 
We do not discuss the details about other techniques except quantization, because the focus of this paper is on quantization.

\section{Weight Normalization based Quantization}\label{sec:wnq}

In this section, we present the detail of our method called weight normalization based quantization~(WNQ). 
WNQ is a multi-bit quantization method. Different from most existing methods which suffer from long-tail weight distribution problem, 
WNQ can effectively avoid this problem by adopting weight normalization.

WNQ quantizes weights in each filter rather than in each layer to achieve better performance.
For a convolution layer $\mathcal{W} \in \mathbb{R}^{N \times C \times s \times s}$, where $N$, $C$ and $s$ represent output channel number, input channel number and kernel size, respectively. 
We reshape each filter $\mathcal{W}_n \in \mathbb{R}^{C \times s \times s}$ to a vector $\mathbf{w}_n \in \mathbb{R}^{Cs^2}$.
For a fully-connected layer $\mathcal{W} \in \mathbb{R}^{N \times C}$, we denote each row as $\mathbf{w}_n \in \mathbb{R}^{C} $.
For both convolution layers and fully-connected layers, we quantize each $\mathbf{w}_n$ separately.
For convenience, in this section, we simply use $\mathbf{w}$ to denote $\mathbf{w}_n$ by omitting the subscript. In addition, we denote the quantized weights of $\mathbf{w}$ as $\mathbf{w}^q$.
Furthermore, we denote the $i$th element in $\mathbf{w}$ as $w_i$ and the number of elements in $\mathbf{w}$ as $M$, that is $\mathbf{w} \in \mathbb{R}^M$. More specifically, $M=Cs^2$ in the convolution layer and $M=C$ in the fully-connected layer.
We use $K$ to denote the bit-width of the quantized weights.

\begin{figure}[t]
  \centering
  \begin{minipage}{.5\textwidth}
    \centering
    \includegraphics[width=0.61\linewidth]{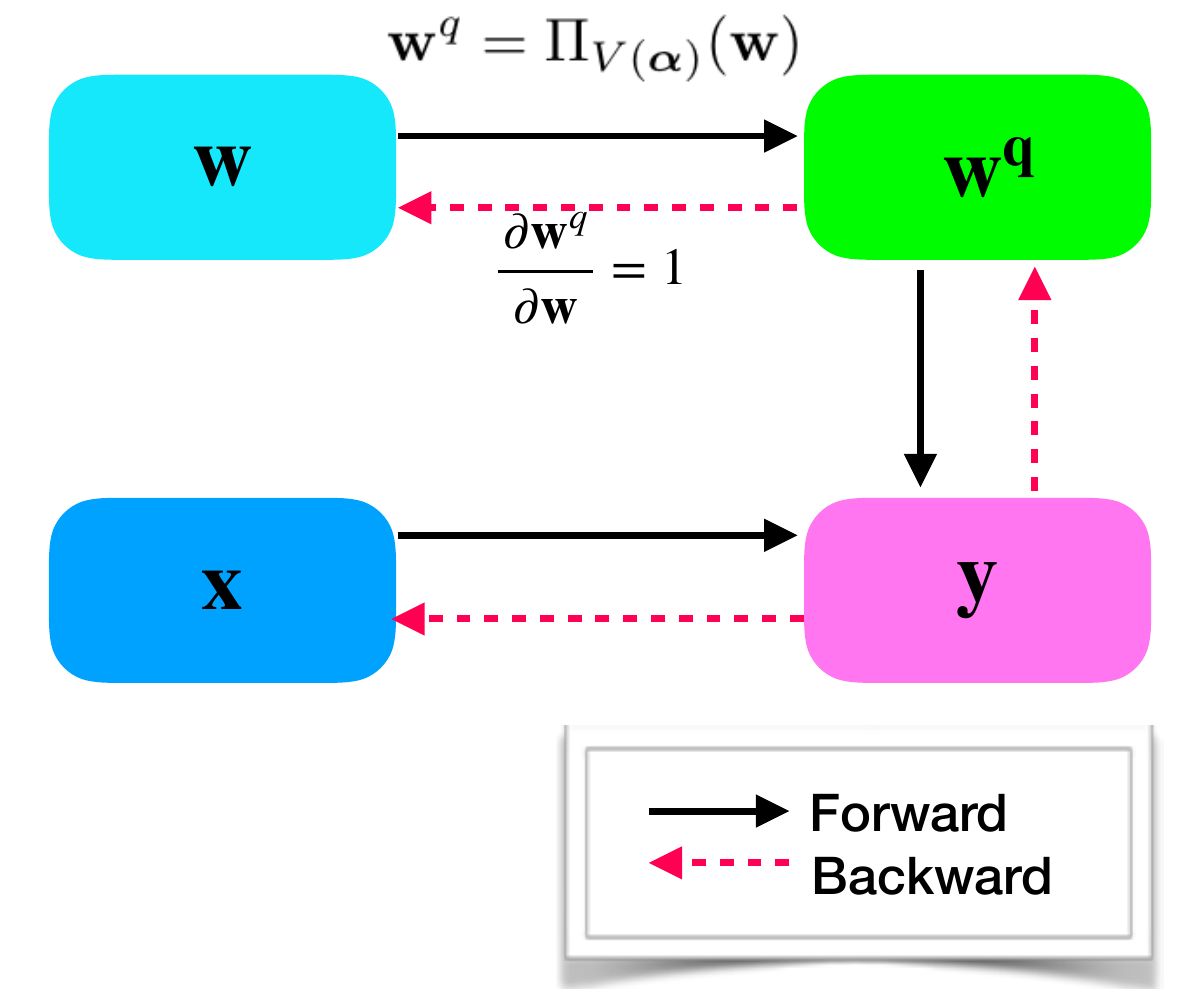}
    \captionof{figure}{One layer in LQ-Net}
    \label{LQ-NET-pro}
  \end{minipage}%
  \begin{minipage}{.5\textwidth}
    \centering
    \includegraphics[width=1.0\linewidth]{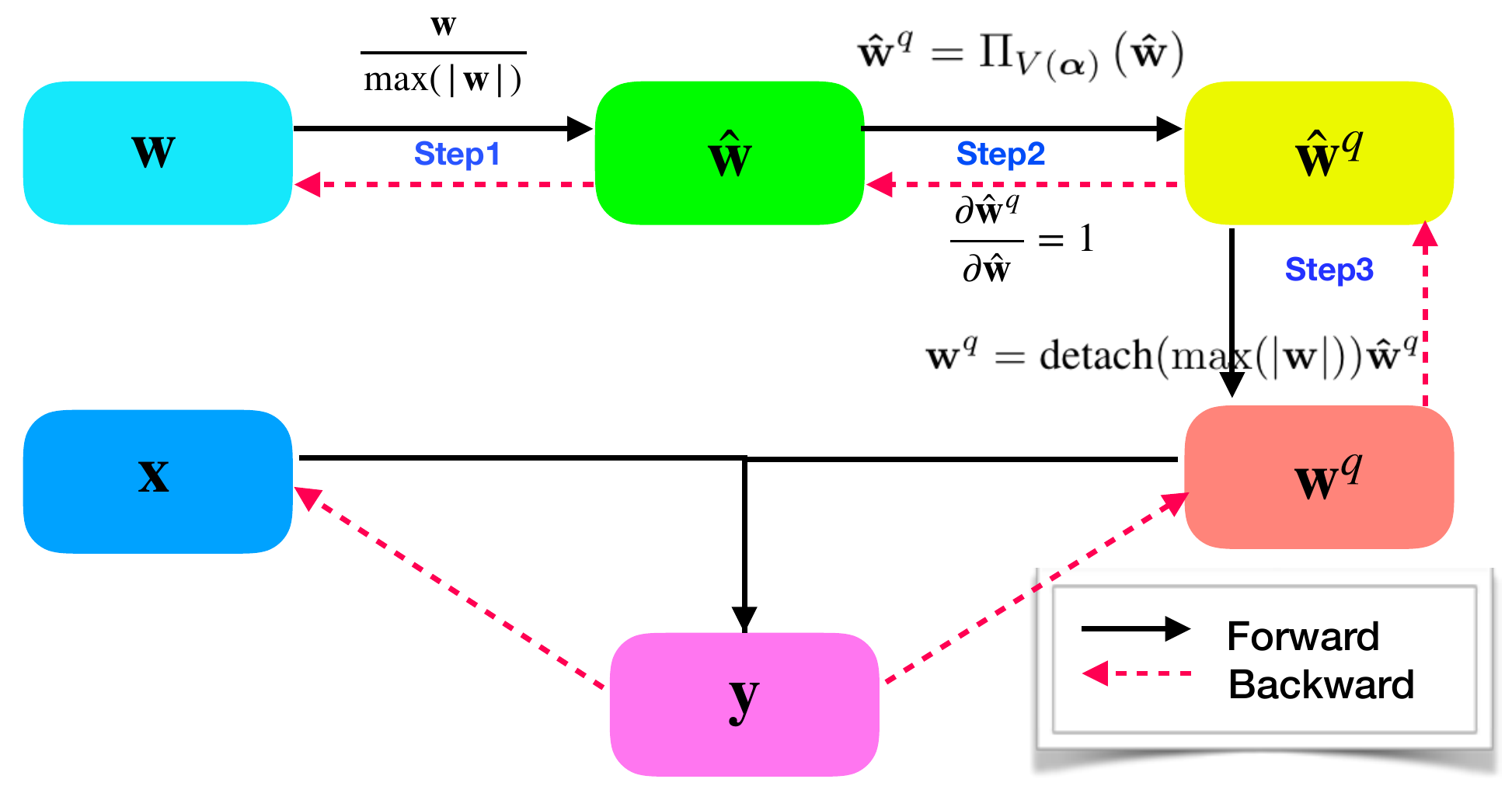}
    \captionof{figure}{One layer in WNQ}
    \label{WNQ-pro}
  \end{minipage}
\end{figure}

The overall forward and backward process of one layer in LQ-Net~\cite{LQ-NET} and WNQ are illustrated in Figure~\ref{LQ-NET-pro} and Figure~\ref{WNQ-pro}, respectively. 
The symbols $\mathbf{x}$ and $\mathbf{y}$ represent the input and output of the corresponding layer. 
In a convolution layer, $\mathbf{y}$ is calculated as follows: $\mathbf{y} = \sigma(\mathbf{w}^q \circledast \mathbf{x})$, where $\sigma$ is the activation function and $\circledast$ is the convolution operation.
The fully-connected layer also has a similar expression.
We can find that WNQ add two extra steps, Step1 and Step3 in Figure~\ref{WNQ-pro}, compared to the process of LQ-Net. 
These two steps are about weight normalization, which is the key to make WNQ avoid the long-tail weight distribution problem.

\subsection{Quantization Function}
\label{Quant_func}
The quantization function is used to map the float weight vector $\mathbf{w}$ to the quantized weight vector $\mathbf{w}^q$.
The whole process of the quantization function is divided into three steps.


\emph{Step1:} 
Firstly, we normalize all weights in the filter to interval $[-1,1]$ through dividing each weight by the maximum absolute value of all weights in this filter:
\begin{align} \label{q_step1}
\mathbf{\hat{w}} = \frac { \mathbf{w}  } { \max \left( \left|  \bf w  \right| \right) },
\end{align}
where the operation $|\cdot|$ means taking absolute value for each element in $\mathbf{w}$, and $\max \left( \left|  \bf w  \right| \right)$ denotes the maximum absolute value~(mav) of all elements in $\mathbf{w}$.

\emph{Step2:} Next, we quantize each element of $\mathbf{\hat w}$ to the nearest quantization level in set $V(\bm \alpha)$.
This quantization step can be explained as the following projection operation $\Pi(\cdot)$:
\begin{align} \label{q_step2}
  \mathbf{\hat{w}} ^q  = \Pi _ {  V(\bm \alpha)  } \left( \mathbf{\hat{w}}  \right).
\end{align}
Here, the projection operation $\Pi(\cdot)$ projects each element in $\mathbf{\hat{w}}$ to the nearest value in the set $V(\bm \alpha)$.
The quantization level set $V(\bm \alpha)$ is parameterized by $\bm \alpha$.
We denote the $l$th binary code as $\mathbf{e}_l \in \{-1,1\}^K $ with $1 \leq l \leq 2^K$,
which enumerates all the binary encodings of $K$-bit from $ [ - 1, \dots, - 1 ] $ to $[1, \dots, 1]$.
The quantization level set $V(\bm \alpha)$ is formulated as  $V(\bm \alpha) = \{\bm {\alpha}^T \mathbf{e}_l | 1 \leq l \leq 2^K\} $.
The parameters $\bm \alpha$ need to be learned. To learn the quantization parameters $\bm \alpha$, we optimize the mean-squared quantization error as that in LQ-Net~\cite{LQ-NET}. The optimization problem is shown as follows:
\begin{align*}
   \min \limits_{\bm \alpha , B } || \mathbf{\hat{w}} - B \bm \alpha||_2^2 \quad s.t. \,\, \bm \alpha \in \mathbb{R}_+^{K}, B \in \{-1,1\}^{M \times K}.
\end{align*}
Alternating optimization is applied to solve this problem. More specifically, each time we optimize one variable by fixing the other one, and this procedure will be repeated for several iterations. To optimize $\bm \alpha$ by fixing $B$, it becomes a regression problem and the solution is as follows:
\begin{align*}
    \bm \alpha^* = (B^T B)^{-1} B^T \mathbf{\hat{w}}.
\end{align*}
To optimize $B$ by fixing $\bm \alpha$, the problem can be solved by projecting each $\hat{w}_i$ to the quantization level set $V(\bm \alpha)$, and then get
the corresponding binary code of the projected quantization level $\mathbf{e}_l$ as a result:
\begin{align*}
    B^*_i = \mathbf{e}_l \quad s.t. \quad  \bm \alpha^T \mathbf{e}_l = \Pi_{V(\bm \alpha)}(\hat{w}_i), \mathbf{e}_l \in \{-1, 1\}^K, 1 \leq l \leq 2^K. 
\end{align*}
By alternatively updating $B$ and $\bm \alpha$ for several iterations, we can get the solution of $\bm \alpha$.
After getting $\bm \alpha$, we use (\ref{q_step2}) to get the quantization of the normalized weights, denoted as $\mathbf{\hat{w}}^q$.

\emph{Step3:}
Note that what we quantize above is the normalized weights. The magnitude of the normalized weights $\mathbf{\hat{w}}$ and that of the original weights $\mathbf{w}$ are different.
What we want to quantize is the original weights $\mathbf{w}$. To get the quantized result of the original weights $\mathbf{w}$,
we multiply the maximum absolute value back to $\mathbf{\hat{w}}^q$ and get $\mathbf{w}^q$:
\begin{align} \label{q_step3}
  \mathbf{w}^q = \text{detach} (\max(|\mathbf{w}|)) \mathbf{\hat{w}}^q.
\end{align}
Here, the detach($\cdot$) operator splits the variable from the computational graph, thus the variable $\text{detach} (\max(|\mathbf{w}|))$ is viewed as a constant here.

The reason why the multiplier is  $\text{detach} (\max(|\mathbf{w}|))$ is easy to show. We can observe that minimizing the original quantization error of $\mathbf{w}$ : $\min \limits_{\bm \alpha , B } || \mathbf{w} - \text{detach}(\max(|\mathbf{w}|)) \cdot B \bm \alpha||_2^2 $  shares the same
solution of $\bm \alpha, B$ with the problem minimizing the quantization error of the normalized weights $\mathbf{\hat{w}}$: $\min \limits_{\bm \alpha , B } || \mathbf{\hat{w}} - B \bm \alpha||_2^2$.
Hence, the original weight quantization $\mathbf{w}^q$ is $\text{detach} (\max(|\mathbf{w}|))$ times the quantization of the normalized weights $\mathbf{\hat{w}}^q$.



\subsection{Back Propagation}

For~(\ref{q_step1}) of \emph{Step1}, the gradient of the function $\max (\cdot)$ is defined as:
\begin{align}
  \frac{\partial \, {\max(|\bf w|)}}{\partial w_i} =
  \begin{cases}
    1 &\mbox{if $w_i$ is the element of maximum absolute value}\\
    0 &\mbox{else}
  \end{cases} \label{g_step1}
\end{align}
Note that the gradient equals to $1$ only when the corresponding element has the maximum absolute value, otherwise the gradient is zero. 
So the gradient of the \emph{Step1} is well defined.

For~(\ref{q_step2}) in \emph{Step2}, it is a step function and the gradient is zero almost everywhere.
As a result, the gradient from the upper layers cannot back propagate to the below layers. To address this problem,
we use straight-through estimator~(STE)~\cite{LQ-NET} to approximate the gradient of~(\ref{q_step2}).
The derivative of $\mathbf{\hat{w}}^q$ to $\mathbf{\hat{w}}$ is set to $1$ everywhere, formulated as $\frac{\partial{\mathbf{\hat{w}}}^q}{\partial{\mathbf{\hat{w}}}} = 1$.

For~(\ref{q_step3}) in \emph{Step3}, there is a detach($\cdot$) operation and this causes the gradient different from~(\ref{g_step1}). The corresponding gradient of~(\ref{q_step3}) is defined as follows:
\begin{align}
  \frac{\partial \, {\text{detach}(max(|\bf w|))}}{\partial w_i} =  0 \quad \text{for every} \, w_i \label{detach_max}
\end{align}
The detach($\cdot$) operation views the variable as a constant, thus the derivative of a constant to every variable is zero. If we do not use detach($\cdot$),
\emph{Step1} and \emph{Step3} are completely opposite operations and getting rid of these two steps does not have any effect on the forward and backward results. Hence, the detach($\cdot$) operation is necessary here.

With the above three separate backward processes, we can define the full back propagation of the WNQ as follows:
\begin{align}
  \frac{\partial L} {\partial w_i} =
  \begin{cases}
     \sum_{j=1}^M \frac{\partial L} {\partial w^q_j} \frac{\partial w^q_j} {\partial w_i}
                                      =  - \sum_{j \neq i}^M  \frac{\partial L} {\partial w^q_j}  \frac{w_j}{w_i}
                                      &  \mbox{if $w_i$ is the element of maximum absolute value} \\
     \frac{\partial L} {\partial w^q_i} & \mbox{else} \label{g_step_all}
  \end{cases}.
\end{align}
Here, $L$ is the loss function (i.e. cross entropy loss) of the network.

\subsection{Discussion}
Here, we discuss the relationship and difference between WNQ and LQ-Net. 
In LQ-Net, only \emph{Step2} is contained but \emph{Step1} and \emph{Step3} do not exist. 
Because quantizing weights with or without \emph{Step1} and \emph{Step3} can get the same quantization weights $\mathbf{w}^q$, hence the forward quantization process are actually equivalent for WNQ and LQ-Net. 
The key difference between WNQ and LQ-Net lies in the back propagation process.
In LQ-Net, the gradient is $\frac{\partial L} {\partial w_i} = \frac{\partial L} {\partial w^q_i}$ for every $w_i$. 
But in WNQ, the gradient of the element with maximum absolute value is different as shown in~(\ref{g_step_all}) and the gradients of other elements in WNQ are equal to those in LQ-Net.

Please note that we perform normalization with the maximum absolute value of the weights, but not a constant. 
If we use a constant for normalization, adding \emph{step1} and \emph{step3} does not have any impact on the forward and backward results, which is equivalent to LQ-Net.

The gradient of the element with maximum absolute value plays an important role in WNQ. 
This gradient is the weighted average of all weight gradients in the filter, except the gradient of the element with maximum absolute value itself. 
This gradient will make the element with maximum absolute value move towards zero in each iteration, which can ultimately avoid~(eliminate) the long-tail of the weight distribution after many iterations. 
After the long-tail of the weight distribution is eliminated, we can approximate the original float weights with quantized weights more accurately. 
This will be empirically verified in Section~\ref{weight_dist}.

\section{Experiments} \label{sec:exp}
We experiment on a cluster with Intel(R) Xeon CPU E5-2620v4@2.1G, 128GB RAM and 4 NVIDIA TITAN XP GPUs.
We implement the proposed method on PyTorch~\cite{pytorch}, 

Two popular datasets, CIFAR-100~\cite{Cifar} and ImageNet~\cite{imagenet} datasets, are adopted for evaluation. 
CIFAR-100 dataset contains 60,000 32x32 color images from 100 different classes, in which 50,000 images are used for training and 10,000 images are used for testing. 
For training, we apply a commonly used data augmentation strategy for preprocessing, which includes padding by 4 pixels on each side, randomly cropping to 32x32, doing a horizontal flip with probability 0.5, 
and normalizing by mean and std.
For testing, we just normalize images by mean and std.

The ImageNet dataset consists of 1.2M/50K training/validation images from 1000 general object classes.
We also apply a commonly used data augmentation strategy for preprocessing, 
which includes resizing the short edge to 256, randomly cropping to 224x224, doing a horizontal flip with probability 0.5, and normalizing by mean and std. 
For testing, we use simple single-crop testing, which resizes the short edge to 256 and center-crop to 224x224.


\subsection{Implementation Details}  \label{hyper-param}
As in~\cite{LQ-NET,HWGQ}, we employ layer re-ordering to the networks for the convenience of integrating activations quantization in the future. 
The classical structure Conv->BN->ReLU-> (Pooling) is replaced by Conv-> (Pooling) ->BN->ReLU, where the pooling layer is optional. 
We quantize all the convolution and fully-connected layers including the first and last layers as in previous methods~\cite{Truct_Gaussian, TEL}. 
By quantizing all layers, we could get a larger speedup rate on hardware than those methods not quantizing the first or last layers.

In all the experiments, quantized model parameters are initialized with a pre-trained full precision model. 
Benefiting from this initialization, we can fine-tune fewer epochs and get less accuracy decreasing than random initialization,
especially when all layers are quantized. 
We optimize all networks with stochastic gradient descent with momentum 0.9. The following hyper-parameters setting is adopted from~\cite{PFGM}.

On CIFAR100, the batch size is 128 and the weight decay is 5e-4.
For the full precision model, the learning rate is initialized to 0.1 and is divided by 5 at the beginning of 60, 120 and 160 epoch, 200 epochs in total. 
For training the quantized models, the learning rate is initialized to 0.01 and divided by 5 at the beginning of 18, 36 and 48 epoch, 60 epochs in total. 

On ImageNet, The batch size is 256 and the weight decay is 1e-4. 
For the full precision model, the learning rate is initialized to 0.1 and is divided by 10 at the beginning of 30, 60, 90 epoch, 100 epochs in total. 
For training the quantized models, the learning rate is initialized to 0.01 and divided by 10 at the beginning of 15, 30, 40 epoch, 45 epochs in total. 

For the quantization parameters $\bm \alpha$, we initialize it with residual quantization~\cite{Residual}. During training, the parameters $\bm \alpha$ in current iteration are initialized by last iteration 
and the alternating optimization in \emph{Step2} in Section~\ref{Quant_func} for updating $\bm \alpha$ is alternated only once in each iteration.


\subsection{CIFAR-100}
As in~\cite{Reg_Act,LQ-NET}, we report the best Top1 test accuracy during the fine-tuning. 
We also report the corresponding Top1 gap between full precision models and quantized models~(the smaller, the better). 
Note that all experiments on CIFAR-100 of previous methods, DoReFa-Net~\cite{Dorefa}, Residual~\cite{Residual} i.e. Sketch (dir.) in the paper and LQ-Net~\cite{LQ-NET}, are re-implemented by ourselves with the same configuration in Section~\ref{hyper-param}, for fair comparisons. 
All experimental results on CIFAR-100 are averaged over five runs.

On CIFAR-100, we use several representative models: 
ResNet20~\cite{ResNet} with its variations PreResNet20~\cite{PreResNet} and SE-ResNet~\cite{SE-NET}\footnote{https://github.com/moskomule/senet.pytorch}, 
Vgg7-wd0.25 and MobileNetv1~\cite{MobileNetv1} \footnote{https://github.com/kuangliu/pytorch-cifar}. 
For ResNet20, we use shortcut typeA~\cite{ResNet}. The comparisons with several baselines are reported in Table~\ref{cifar-resnet20}. 
We can find that, with the same bit-width, WNQ outperforms other methods to achieve the best performance.
In particular, WNQ outperforms the LQ-Net by 0.89\% in testing error in 2-bit weights~(i.e. 16x compression rate). 
WNQ with 4-bit can even outperform the full precision model by 0.06\%.
The test accuracy curve of ResNet20 is illustrated in Figure~\ref{resnet20_curve}.
\begin{wraptable}{R}{75mm}  
  \caption{ResNet20 on CIFAR-100. The smaller the Top1-gap, the better. FP denotes the full precision model. (in percent)}
  \label{cifar-resnet20}
  \centering
  \begin{tabular}{lccc}
    \toprule
    \multicolumn{4}{c}{ResNet20}                   \\
    \cmidrule(r){1-4}
    Method     & bit    & Top1 & Top1-gap \\
    \hline
    DoReFa-Net                    & $2$    & $66.65$     & $2.33$ \\
    Residual                      & $2$    & $65.97$     & $3.01$ \\
    LQ-Net                        & $2$    & $66.53$     & $2.45$ \\
    WNQ                           & $2$    & $67.42$     & $\bf 1.56$ \\
    \hline
    DoReFa-Net                    & $3$    & $68.40$     & $0.58$ \\
    Residual                      & $3$    & $68.29$     & $0.69$ \\
    LQ-Net                        & $3$    & $68.39$     & $0.59$ \\
    WNQ                           & $3$    & $68.80$     & $\bf 0.18$ \\
    \hline
    DoReFa-Net                    & $4$    & $68.89$     & $0.09$ \\
    Residual                      & $4$    & $68.70$     & $0.28$ \\
    LQ-Net                        & $4$    & $69.01$     & -$0.03$ \\
    WNQ                           & $4$    & $69.04$     & -$\bf 0.06$ \\
    \hline
    FP                            & $32$   & $68.98$     & $-$  \\
    \bottomrule   
  \end{tabular}
\end{wraptable}

For other networks on CIFAR-100, the results are reported in Table~\ref{cifar-commonnet}. 
The Vgg7-wd0.25 is based on Vgg7~\cite{TBN}. We cut the channel number to 0.25x of the original model because the original model gets no significant accuracy drop with only 1-bit. 
We don't quantize the SE-modules of SE-ResNet. We also don't quantize the depth-wise convolution layers of MobileNetv1 as in~\cite{HBNN}. 
Because the number of parameters in these two modules are very small compared with other modules. 
Because 3-bit is enough to preserve the performance of the full precision model, the gaps of all methods compared with the full precision model are small. 
So the improvement of WNQ over LQ-Net in 3-bit is considerable. 
In particular, on ResNet20 along with its variations PreResNet20 and SE-ResNet20, the performance of WNQ is much better than that of LQ-Net. 
Overall, WNQ could achieve the best results in most cases.



\begin{table}[h] 
  \caption{Various networks on CIFAR-100. 2w and 3w mean 2-bit weights and 3-bits weights, respectively. 
  ``gap'' denotes Top1-gap and a smaller number of gap is better. (in percent)}
  \label{cifar-commonnet}
  \begin{tabular}{l|l|c|c|c|c|c|c|c|c}
  \toprule
  \multicolumn{2}{c|}{\multirow{2}{*}{Networks}} & \multicolumn{2}{c|}{PreResNet20} & \multicolumn{2}{c|}{SE-ResNet20} & \multicolumn{2}{c|}{Vgg7-wd0.25} & \multicolumn{2}{c}{MobileNetv1} \\ \cline{3-10} 
  \multicolumn{2}{c|}{}                  & \multicolumn{2}{c|}{FP:68.65}    & \multicolumn{2}{c|}{FP:69.30}     & \multicolumn{2}{c|}{FP:67.30}       & \multicolumn{2}{c}{FP:68.96}       \\ \hline
  \multirow{4}{*}{2w}            &                           & Top1           & gap           & Top1           & gap         & Top1           & gap          & Top1                  & gap        \\ \cline{2-10} 
                                 &DoReFa-Net                 & $65.71$        & $2.94$        & $66.75$        & $2.55$      & $66.41$        & $0.89$       & $52.17$               & $16.79$      \\ \cline{2-10} 
                                 &LQ-Net                     & $65.75$        & $2.90$        & $66.71$        & $2.59$      & $66.26$        & $1.04$       & $67.50$               & $1.46$      \\ \cline{2-10} 
                                 &WNQ                        & $66.39$        & $\bf 2.26$    & $67.78$        & $\bf 1.52$  & $66.60$        & $\bf 0.70$   & $67.91$               & $\bf 1.05$      \\ \hline \hline
  \multirow{4}{*}{3w}            &                           & Top1           & gap           & Top1           & gap         & Top1           & gap          & Top1                  & gap      \\ \cline{2-10} 
                                 &DoReFa-Net                 & $68.12$        & $0.53$        & $68.55$        & $0.75$      & $66.91$        & $0.39$       & $67.58$               & $1.38$      \\ \cline{2-10} 
                                 &LQ-Net                     & $68.29$        & $\bf 0.36$    & $69.11$        & $0.19$      & $66.93$        & $0.37$       & $68.69$               & $0.27$      \\ \cline{2-10} 
                                 &WNQ                        & $68.15$        & $0.50$        & $69.25$        & $\bf 0.05$  & $67.04$        & $\bf 0.26$   & $68.91$               & $\bf 0.05$    \\ \bottomrule
  \end{tabular}
\end{table}

\subsection{ImageNet}
We further apply our method to two representative networks on ImageNet: ResNet \cite{ResNet} and MobileNetv1 \cite{MobileNetv1}. 
We report the best Top5 accuracy during the fine-tuning and also the Top1 accuracy corresponding to the best Top5 model.
Note that the accuracy of full precision models by our implementation may be a little lower because we do not use strong data augmentation as in~\cite{LQ-NET}.
\begin{table}
  \caption{ResNet18 and ResNet50 on ImageNets. T refers to ternary which also needs 2-bit. FP refers to the full precision model. 
  The smaller the Top1-gap and the Top5-gap, the better.
  (in percent)}
  \label{resnet-imagenet}
  \begin{threeparttable}
  \centering
  \begin{tabular}{lc|ccc|ccc}
  \toprule
  \multicolumn{8}{c}{ResNet18}                   \\
  \hline
    Method                                       & bit   & FP Top1 & Top1 & Top1-gap & FP Top5 & Top5 & Top5-gap \\
    \hline
    TTQ \tnote{a}~\cite{TTQ}                     & T      & $69.60$ & $66.60$     & $3.00$ & $89.20$     & $87.20$ & $2.00$ \\
    ADMM \tnote{a} \cite{ADMM}                   & T      & $69.10$ & $67.00$     & $2.10$ & $89.00$     & $87.50$ & $1.50$ \\
    LR-NET \tnote{b} \cite{LR-NET}               & T      & $69.57$ & $63.50$     & $6.07$ & $89.24$     & $84.80$ & $4.44$ \\
    T-Gaussian \tnote{c} \cite{Truct_Gaussian}   & T      & $69.75$ & $66.01$     & $3.74$ & $89.07$     & $86.78$ & $2.29$ \\
    DoReFa-Net \tnote{*} \tnote{c} \cite{Dorefa} & $2$    & $69.22$ & $66.29$     & $2.93$ & $88.84$     & $86.97$ & $1.87$ \\
    ABC \tnote{a} \cite{ABC}                     & $2$    & $69.30$ & $63.70$     & $5.60$ & $89.20$     & $85.20$ & $4.00$ \\
    LQ-Net \tnote{*} \tnote{c} \cite{LQ-NET}     & $2$    & $69.22$ & $67.20$     & $2.02$ & $88.84$     & $87.59$ & $1.25$ \\
    QIL \tnote{a} \cite{QIL}                     & $2$    & $70.20$ & $68.10$     & $2.10$ & $89.60$     & $88.30$ & $1.30$ \\
    WNQ \tnote{c}                                & $2$    & $69.22$ & $67.71$     & $\bf 1.51$ & $88.84$ & $87.86$ & $\bf 0.98$ \\
    \hline
    DoReFa-Net \tnote{*} \tnote{c} \cite{Dorefa}& $3$    & $69.22$ & $67.79$     & $1.43$ & $88.84$   & $87.91$     & $0.93$ \\
    Sketch (ref.) \tnote{a} \cite{Residual}     & $3$    & $68.80$ & $67.80$     & $1.00$ & $89.00$   & $88.40$     & $0.60$ \\
    ABC \tnote{a} \cite{ABC}                    & $3$    & $69.30$ & $66.20$     & $3.10$ & $89.20$   & $86.70$     & $2.50$ \\
    ADMM \tnote{a} \cite{ADMM}                  & $3$    & $69.10$ & $68.00$     & $1.10$ & $89.00$   & $88.30$     & $0.70$ \\
    LQ-Net \tnote{*} \tnote{c} \cite{LQ-NET}    & $3$    & $69.22$ & $68.56$     & $0.66$ & $88.84$   & $88.46$     & $0.38$ \\
    QIL \tnote{a} \cite{QIL}                    & $3$    & $70.20$ & $69.90$     & $\bf 0.30$ & $89.60$   & $89.30$     & $0.30$ \\
    WNQ \tnote{c}                               & $3$    & $69.22$ & $68.83$     & $0.39$ & $88.84$   & $88.60$ & $\bf 0.24$ \\
    \hline
    \hline
    \multicolumn{8}{c}{ResNet50} \\ 
    \hline
    T-Gaussian \tnote{c} \cite{Truct_Gaussian}   & T      & $76.13$ & $73.97$     & $2.16$     & $92.86$   & $91.65$     & $1.21$ \\
    ADMM \tnote{a} \cite{ADMM}                   & $2$    & $75.30$ & $72.50$     & $2.80$     & $92.20$   & $90.70$     & $1.50$ \\
    LQ-Net \tnote{*} \tnote{c} \cite{LQ-NET}     & $2$    & $74.79$ & $73.83$     & $0.96$     & $92.09$   & $91.81$     & $0.28$  \\
    WNQ \tnote{c}                                & $2$    & $74.79$ & $74.07$     & $\bf 0.72$ & $92.09$   & $92.06$     & $\bf 0.03$    \\
  \bottomrule   
\end{tabular}
\begin{tablenotes}
  \item[*] denotes our implementation with configuration in Section~\ref{hyper-param}. Other results without \tnote{*} are quoted from corresponding papers. 
  \item[a] denotes quantizing all layers except the first layer and the last layer.
  \item[b] denotes quantizing all layers except the last layer.
  \item[c] denotes quantizing all layers.
\end{tablenotes}
\end{threeparttable}
\end{table}

\begin{figure}
  \centering
  \begin{minipage}{.48\textwidth}
    \centering
    \includegraphics[width=1.0\linewidth]{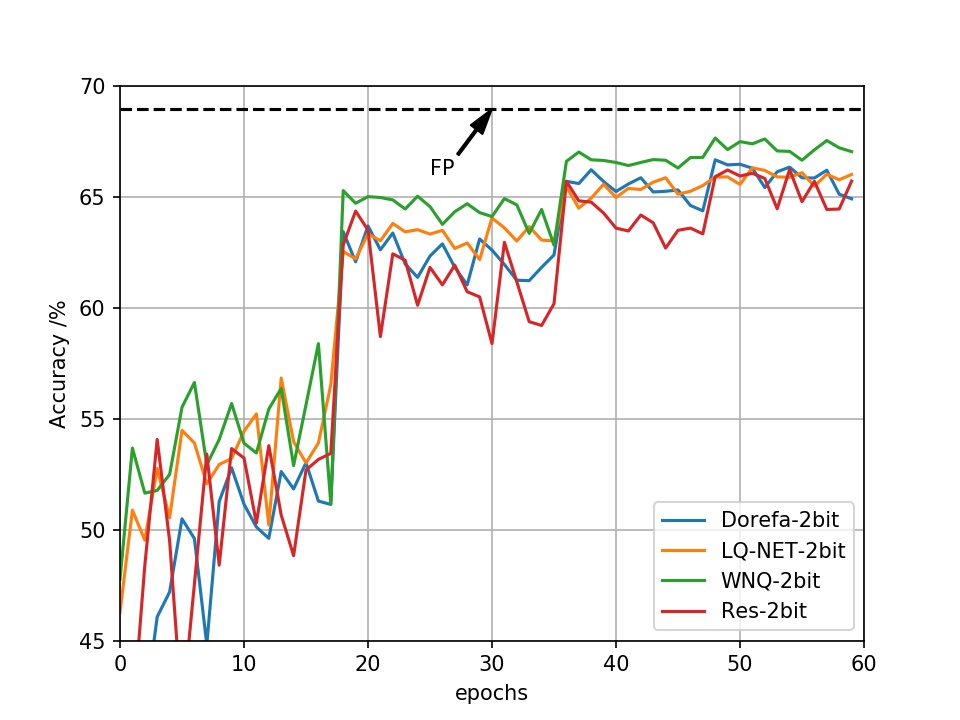}
    \caption{Top1 accuracy of ResNet20 on CIFAR-100.}
    \label{resnet20_curve}
  \end{minipage}%
  \hfill
  \begin{minipage}{.48\textwidth}
    \centering
    \includegraphics[width=1.0\linewidth]{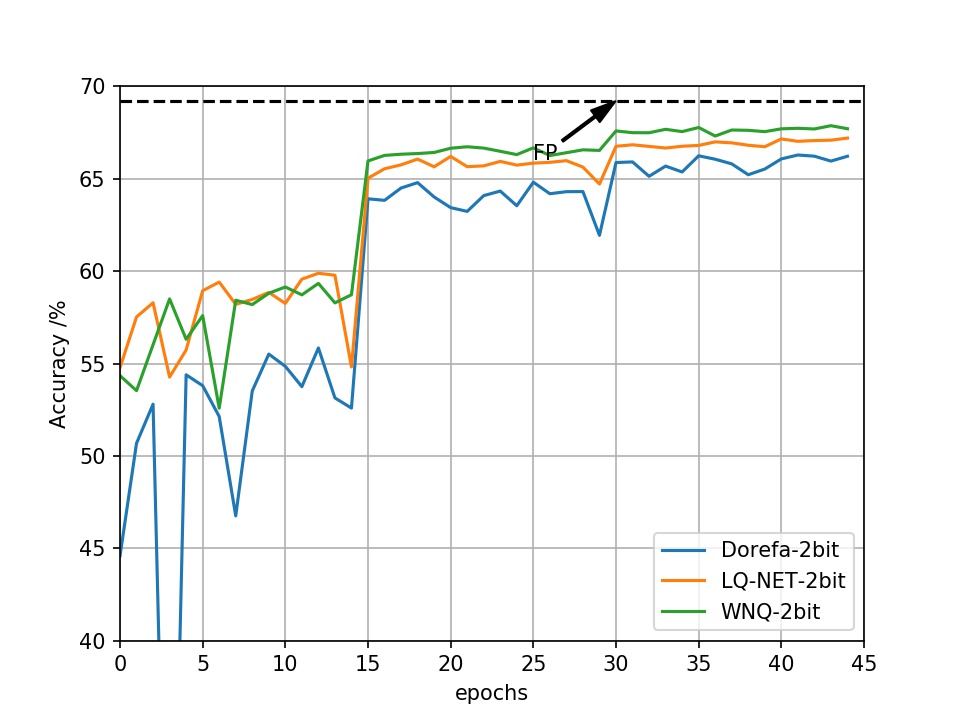}
    \caption{Top1 accuracy of ResNet18 on ImageNet.}
    \label{resnet18_curve}
  \end{minipage}%
\end{figure}

For ResNet18 and ResNet50, we use shortcut typeB~\cite{ResNet}. Results are reported in Table~\ref{resnet-imagenet}.
We can find that WNQ can achieve the best results in most cases compared with the state-of-the-art methods, 
even though we quantize all layers and most of the existing methods do not quantize the first layer and the last layer.
Note that WNQ cannot outperform QIL in terms of the Top1-gap with 3-bit, the main reason is that QIL \cite{QIL} does not quantize the first and the last layer. 
Without quantizing the first and the last layer, performance improvement can be expected for our WNQ.
The validation accuracy curve of ResNet18 is illustrated in Figure~\ref{resnet18_curve}. 

For MobileNetv1, results are reported in Table~\ref{mobilenet-imagenet}. We do not quantize depth-wise convolution layers as in CIFAR-100. 
Our method outperforms LQ-Net~\cite{LQ-NET} by 1.24\% and 1.01\% in Top1 and Top5 accuracy with 2-bit quantization.
Moreover, WNQ is superior to LQ-Net\cite{LQ-NET} by 2.06\% and 1.53\% in Top1 and Top5 accuracy with 3-bit quantization.

\begin{table} 
  \caption{MobileNetv1 on ImageNet. LQ-Net is our implementation. (in percent)}
  \label{mobilenet-imagenet}
  \centering
  \begin{tabular}{lc|ccc|ccc}
    \toprule
    Method     & bit    & FP Top1 & Top1 & Top1-gap & FP Top5 & Top5 & Top5-gap \\
    \hline
    \hline
    LQ-Net    & $2$    & $68.80$ & $62.17$     & $6.63$      &$88.66$ & $84.42$     & $4.24$\\
    WNQ       & $2$    & $68.80$ & $63.41$     & $\bf 5.39$  &$88.66$ & $85.43$    & $\bf 3.23$\\
    \hline
    LQ-Net    & $3$    & $68.80$ & $65.03$     & $3.77$      &$88.66$ &  $86.29$    & $2.37$\\
    WNQ       & $3$    & $68.80$ & $67.09$     & $\bf 1.71$  &$88.66$ &  $87.82$    & $\bf 0.84$ \\
    \bottomrule   
  \end{tabular}
\end{table}

\subsection{Weights Distribution} \label{weight_dist}
\begin{figure}
  \centering
  \includegraphics[width=0.95\textwidth]{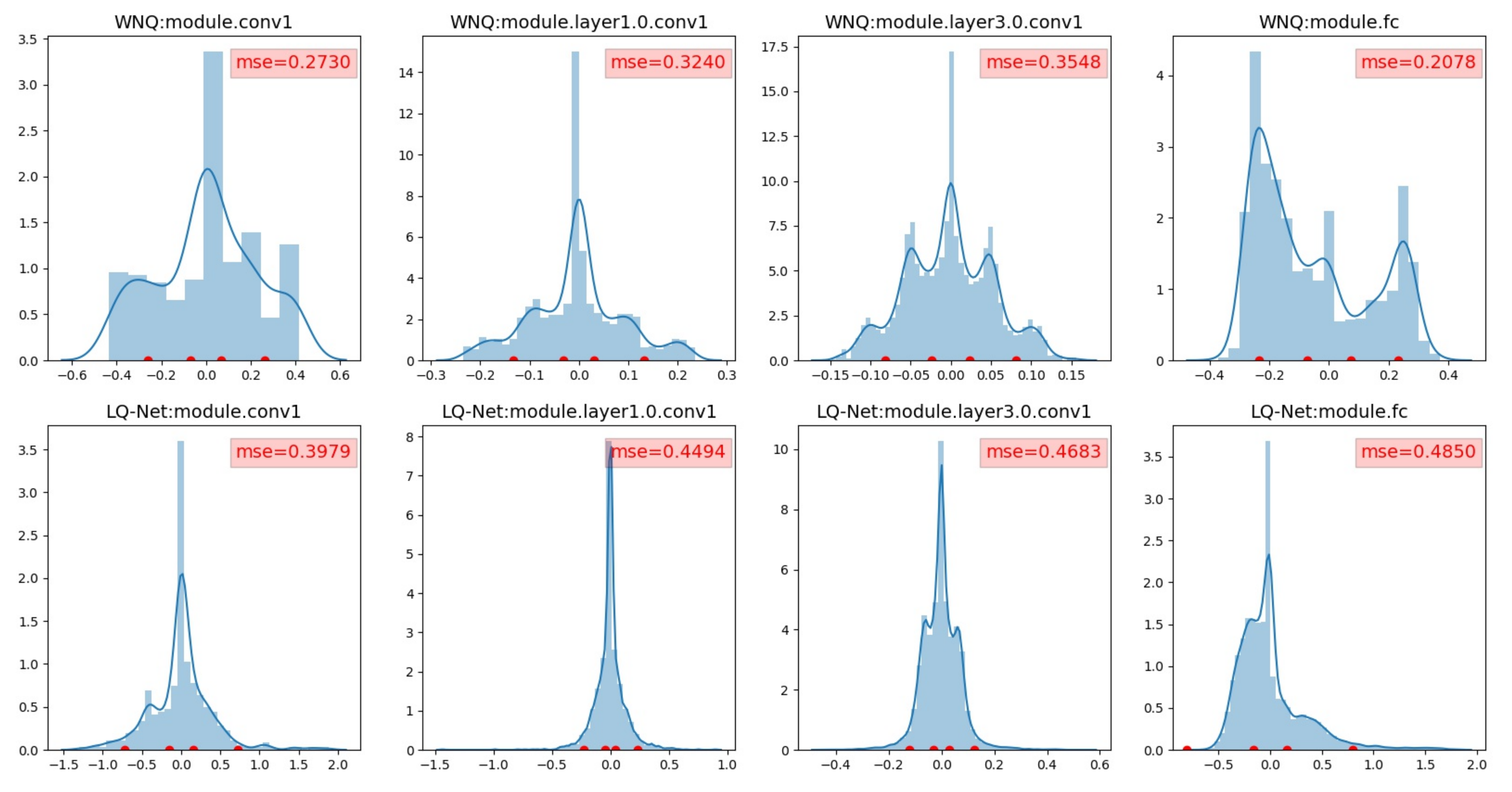}
  \caption{Distribution of float weights $\mathbf{w}$ on some selected layers of ResNet20 on CIFAR-100 in 2-bit setting. Top row is WNQ and bottom row is LQ-Net. 
   Red dots on x-axis are the average quantization levels in this layer. 
   ``mse'' in each figure denotes the relative mean-squared quantization error of the layer defined in Section~\ref{weight_dist}.}
  \label{dist_all}
\end{figure}

In this section, we study the distribution of weights in WNQ and LQ-Net to explain the advantages of WNQ. 
As shown from the Figure~\ref{dist_all}, we can see the maximum absolute value of the weights in WNQ is much smaller than that of LQ-Net.
Furthermore, the distribution of LQ-Net has a long-tail, while the distribution of our WNQ is much smooth and the tail is shorter than that of LQ-Net.
The long-tail distribution will cause a larger quantization error which is denoted as ``mse''~(relative mean-squared quantization error) in Figure~\ref{dist_all}.
The ``mse'' is defined as: $\frac{1}{N} \sum_{n=1}^N \frac{||\mathbf{w}_n - \mathbf{w}_n^q||_2^2}{||\mathbf{w}_n||_2^2}$, which averages over the realtive mean-squared quantization error of each filter.
In particular, for the last fully-connected layer, our WNQ can achieve $5.29\times$ less ``mse'' than LQ-Net.
With smaller relative quantization error, WNQ can achieve better performance on classification accuracy.


\section{Conclusion} \label{sec:cons}
In this paper, we propose a new quantization method, called WNQ, for deep neural network compression. 
WNQ adopts weight normalization to avoid the long-tail distribution of network weights, and subsequently reduces the quantization error.  
Experiments on CIFAR-100 and ImageNet show that WNQ can outperform other baselines to achieve state-of-the-art performance. 
To the best of our knowledge, this is the first work to study the effect of weight distribution for quantization. 
In our future work, we will apply this idea to activation quantization.

\medskip

\small

\bibliographystyle{plainnat}

\bibliography{ref}

\end{document}